\documentclass[sigconf]{acmart}
\usepackage{enumitem}
\usepackage{multirow}
\usepackage{colortbl} 
\usepackage{xcolor}
\usepackage[normalem]{ulem}
\useunder{\uline}{\ul}{}
\AtBeginDocument{%
  }

\setcopyright{acmlicensed}
\copyrightyear{2026}
\acmYear{2026}
\setcopyright{cc}
\setcctype{by}
\acmConference[WWW '26]{Proceedings of the ACM Web Conference 2026}{April 13--17, 2026}{Dubai, United Arab Emirates}
\acmBooktitle{Proceedings of the ACM Web Conference 2026 (WWW '26), April 13--17, 2026, Dubai, United Arab Emirates}
\acmPrice{}
\acmDOI{10.1145/3774904.3792383}
\acmISBN{979-8-4007-2307-0/2026/04}
\begin{document}
\title{Re-understanding Graph Unlearning through Memorization}

\author{Pengfei Ding}
\affiliation{%
 \institution{Macquarie University}
 \city{Sydney}
 \country{Australia}}
\email{pengfei.ding@mq.edu.au}

\author{Yan Wang}
\authornote{Corresponding author}
\affiliation{%
 \institution{Macquarie University}
 \city{Sydney}
 \country{Australia}}
\email{yan.wang@mq.edu.au}

\author{Guanfeng Liu}
\affiliation{%
 \institution{Macquarie University}
 \city{Sydney}
 \country{Australia}}
\email{guanfeng.liu@mq.edu.au}

\begin{abstract}
Graph unlearning (GU), which removes nodes, edges, or features from trained graph neural networks (GNNs), is crucial in Web applications where graph data may contain sensitive, mislabeled, or malicious information. However, existing GU methods lack a clear understanding of the key factors that determine unlearning effectiveness, leading to three fundamental limitations: (1) impractical and inaccurate GU difficulty assessment due to test-access requirements and invalid assumptions, (2) ineffectiveness on hard-to-unlearn tasks, and (3) misaligned evaluation protocols that overemphasize easy tasks and fail to capture true forgetting capability. To address these issues, we establish GNN memorization as a new perspective for understanding graph unlearning and propose MGU, a Memorization-guided Graph Unlearning framework. MGU achieves three key advances: it provides accurate and practical difficulty assessment across different GU tasks, develops an adaptive strategy that dynamically adjusts unlearning objectives based on difficulty levels, and establishes a comprehensive evaluation protocol that aligns with practical requirements. Extensive experiments on ten real-world graphs demonstrate that MGU consistently outperforms state-of-the-art baselines in forgetting quality, computational efficiency, and utility preservation.
\end{abstract}



\maketitle

\section{Introduction}
The widespread use of Graph Neural Networks (GNNs) in Web-related domains (e.g., social networks \cite{guo2020deep}, recommender systems \cite{sharma2024survey}, financial applications \cite{wang2021review}) poses unprecedented opportunities and challenges. In the Web ecosystem, graph-structured data inevitably includes sensitive, erroneous, or malicious information (e.g., privacy-sensitive records, mislabeled interactions, fraudulent transactions \cite{wang2023inductive}), which not only degrades the reliability of Web services but also raises critical privacy and security concerns, highlighting the necessity of removing such information from trained GNNs. Graph Unlearning (GU) \cite{fan2025opengu, ding2025adaptive} has thus emerged as a crucial paradigm for the Web, enabling efficient removal of specific graph elements (nodes, edges, or features) without costly full retraining.

Despite growing research interest, the key factors that determine graph unlearning effectiveness remain unclear. Understanding what makes an unlearning task easy or hard is crucial, as such insights can guide the development of unlearning methods that are robust against challenging cases, and support evaluation protocols aligned with practical requirements~\cite{zhao2024makes}. Motivated by this, we investigate whether existing methods can effectively assess graph unlearning difficulty. However, our study shows that current difficulty assessments are both impractical and inaccurate, which not only masks the ineffectiveness of existing methods on hard-to-unlearn tasks, but also leads to misaligned evaluation protocols. Specifically, we identify the following three interconnected limitations:

\begin{figure}[t]
\centering
\scalebox{0.392}{\includegraphics{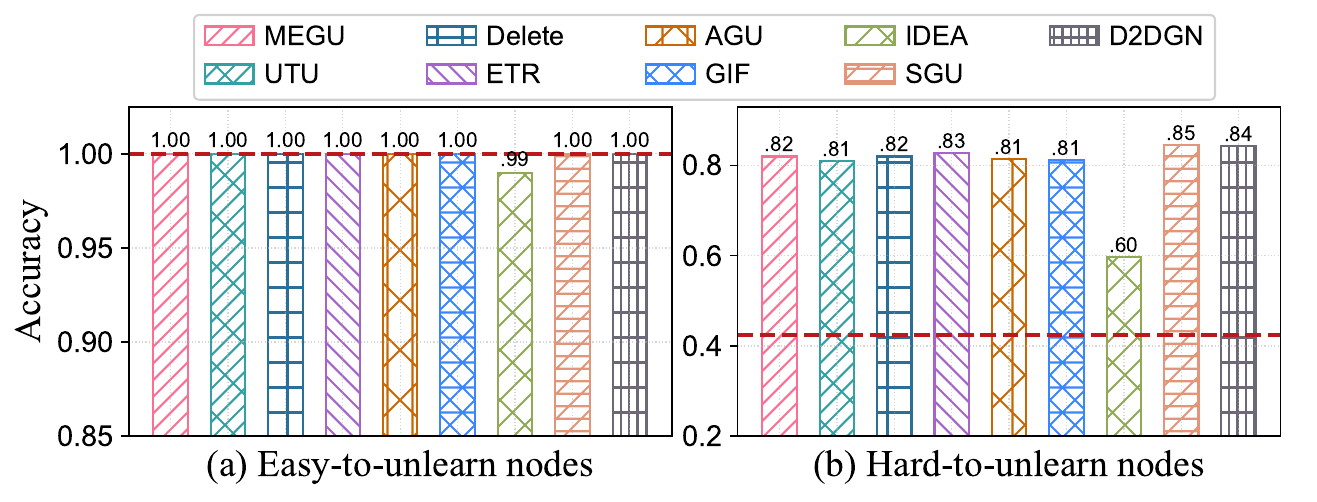}}
\caption{Unlearning performance on the 10\% easiest and 10\% hardest nodes in the Cora dataset. The red dashed line denotes the gold-standard performance obtained by retraining.}
\Description{A diagram showing the effect of deleting a node or edge in a graph on different graph neural networks.}
\label{fig1}
\end{figure}

\textbf{Limitation 1: Impractical and inaccurate difficulty assessment.} 
Existing assessment strategies \cite{cheng2023gnndelete, chen2025frog} suffer from two key issues: (1) they require access to test data or already-unlearned models, which are typically unavailable in practice; and (2) they assume that unlearning difficulty depends on graph connectivity distance to the test set. However, our analysis reveals that unlearning difficulty shows no correlation with test-set distance (see Section \ref{4.2}).

\textbf{Limitation 2: Ineffectiveness on hard-to-unlearn tasks.}
Our difficulty assessment reveals a critical blind spot: existing GU methods perform well on easy-to-unlearn tasks but fail dramatically on hard ones. Figure \ref{fig1} illustrates this gap on the Cora dataset with a GCN backbone \cite{kipf2016semi}, where we compare unlearning performance on the 10\% easiest and 10\% hardest training nodes (ranked by our metric) and evaluate forgetting quality by re-inputting deleted nodes into the unlearned model. Ideally, effective unlearning should yield predictions consistent with the \textit{gold standard}, i.e., a model retrained from scratch without the deleted nodes (red dashed line in Figure \ref{fig1}). While all methods achieve near-optimal results on easy-to-unlearn nodes, their performance drops by over 40\% on hard ones, as unlearned models continue predicting the deleted nodes' labels with high accuracy, rather than exhibiting the retrained model's uncertainty. This degradation consistently appears across multiple datasets, GNN backbones, and unlearning tasks (see Appendix), indicating a systematic limitation of existing GU methods.

\begin{figure}[t]
\centering
\scalebox{0.396}{\includegraphics{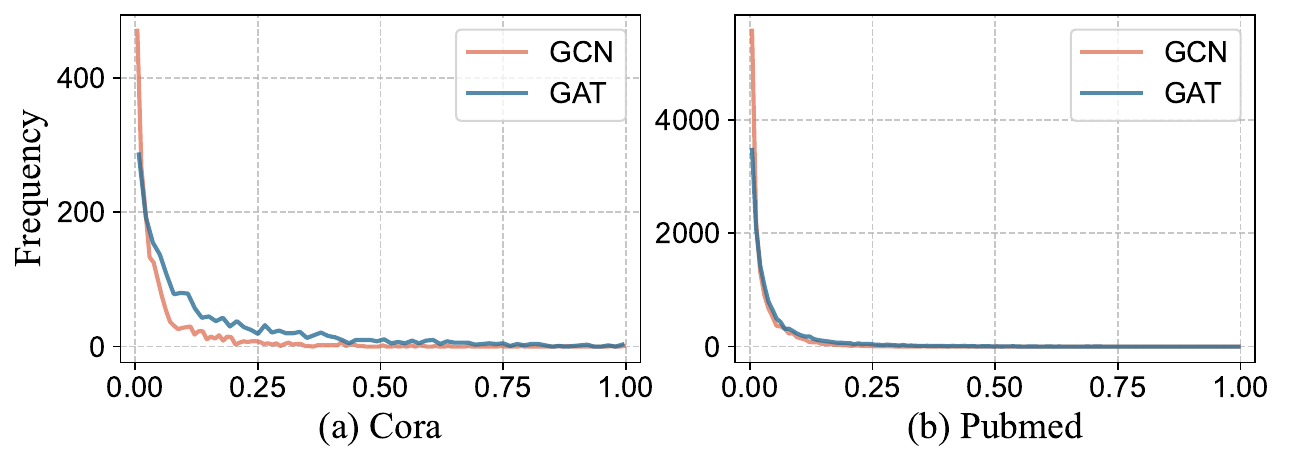}}
\caption{Long-tailed distribution of node unlearning difficulty (via Eq.~(\ref{eq2})). Higher scores indicate harder cases.}
\Description{A diagram showing the effect of deleting a node or edge in a graph on different graph neural networks.}
\label{eg2}
\end{figure}

\textbf{Limitation 3: Misaligned evaluation protocols.} The above limitations reveal a misalignment between current evaluation protocols and practical needs in two key aspects: (1) \textbf{Selection bias:} Most evaluations select unlearning targets randomly \cite{ding2025adaptive, wu2023gif}, which fails to reflect realistic unlearning scenarios. As shown in Figure \ref{eg2}, unlearning difficulty follows a long-tailed distribution, where random sampling tends to favor easy-to-unlearn elements and overlooks hard ones. However, these hard-to-unlearn elements, though rare and structurally irregular, often correspond to actual unlearning targets in practice (e.g., privacy-sensitive outliers, mislabeled data, malicious accounts \cite{zhao2024makes, feldman2020what}).
(2) \textbf{Insufficient verification:} Current evaluations focus on utility preservation but rarely verify whether deleted information is fully removed \cite{yang2024erase,li2024towards}. Some methods assess forgetting on deleted elements \cite{fan2025opengu} but assume uniform degradation in predictions (e.g., convergence to random guessing), ignoring that elements at different difficulty levels exhibit distinct forgetting behaviors. As shown in Figure \ref{fig1}, gold-standard retraining produces different patterns for easy and hard tasks, illustrating how this uniform assumption can mislead the design of GU methods.

These limitations reveal that existing studies understand GU from superficial perspectives, emphasizing structural effects or empirical observations while overlooking the fundamental insight: \textit{the way a model memorizes information inherently determines how it should be forgotten}. \textit{Memorization} characterizes how a model encodes and retains information during training, fundamentally governing the persistence of learned knowledge and the difficulty of removing it \cite{feldman2020what, wei2024memorization}. By exploring how GNNs memorize graph elements to varying degrees, we can better understand why some elements are harder to unlearn and how to handle them effectively. 


\textbf{Our Work.} In this paper, we establish GNN memorization as a new perspective for understanding GU, and propose MGU, a \textbf{\underline{M}}emorization-guided \textbf{\underline{G}}raph \textbf{\underline{U}}nlearning framework. To address \textbf{Limitation 1}, we establish the relationship between memorization and unlearning difficulty, and design an accurate assessment framework without test-access requirement. To address \textbf{Limitation 2}, we propose an adaptive unlearning strategy that adjusts objectives based on how unlearning difficulty affects generalization. To address \textbf{Limitation 3}, we propose a novel evaluation protocol with difficulty-aware sampling and comprehensive performance assessment that more faithfully reflects real-world requirements.

Our main contributions are summarized as follows:
\begin{itemize} [leftmargin=*]
\item \textbf{Novel Perspective}: We establish the first connection between GNN memorization and unlearning difficulty, providing new insights into why existing methods fail on hard-to-unlearn tasks.

\item \textbf{Universal Adaptive Framework}: We propose a general model-agnostic unlearning framework that adaptively handles both easy and hard GU tasks without accessing model's internal structure.

\item \textbf{Comprehensive Evaluation Protocol}: We design a difficulty-aware sampling strategy with comprehensive performance assessment that better aligns with real-world requirements.

\item \textbf{SOTA Performance}: Extensive experiments on ten real-world graphs demonstrate the superiority of MGU in forgetting quality, computational efficiency, and utility preservation.
\end{itemize}

\section{Related Work}
\subsection{Graph Unlearning}
Early research on GU primarily focused on specific tasks or GNN architectures: (1) Partition-based methods \cite{chen2022graph,wang2023inductive,zhang2024revoke} partition the graph into shards and retrain only the affected submodels to reduce computational cost. However, these methods are restricted to node unlearning and become inefficient on large-scale graphs. (2) Closed-form methods \cite{cong2022grapheditor,chien2023efficient,yi2024scalable} enable retraining-free updates with theoretical guarantees, but rely on stringent assumptions about linear GNNs and lack generalizability. 

To overcome these limitations, recent studies \cite{li2024tcgu,yang2024erase} propose general frameworks for diverse GU tasks and GNN backbones. (1) Influence function-based methods \cite{wu2023gif,dong2024idea} approximate parameter updates for efficient unlearning across tasks. (2) Learning-based methods \cite{cheng2023gnndelete,li2024towards,ding2025adaptive} design specialized loss functions to balance forgetting and reasoning. However, these methods remain effective on easy-to-unlearn tasks and degrade significantly on hard ones.

\subsection{Memorization in Deep Neural Networks}
In Euclidean domains such as images and text, memorization in deep neural networks (DNNs) has been extensively studied \cite{arpit2017closer,wei2024memorization}. Recent theories \cite{yu2024generalizablity} show that memorization is essential for near-optimal generalization. Empirical studies \cite{zhao2024makes} further reveal that atypical or noisy samples are more likely to be memorized.


Despite extensive progress in Euclidean domains, studies of memorization in GNNs remain limited, and its connection to unlearning difficulty is unexplored. A recent work \cite{jamadandi2025memorization} extended memorization analysis to GNNs but ignored graph connectivity, focusing only on nodes without considering other graph elements (edges and features). In this work, we present the first systematic study of GNN memorization and reveal its key implications for graph unlearning.

\section{Preliminary}
\subsection{Notations and Background}
In general, a graph is represented as $\mathcal{G}=(\mathcal{V}, \mathcal{E}, \mathcal{X})$, where $\mathcal{V}$ is the set of nodes, $\mathcal{E}$ is the set of edges, and $\mathcal{X}=\{\mathbf{x}_1, \mathbf{x}_2, \dots, \mathbf{x}_n\}$ denotes the node feature matrix. Each node $v_i \in \mathcal{V}$ is associated with a $d$-dimensional feature vector $\mathbf{x}_i \in \mathbb{R}^d$. In this paper, we focus on the commonly studied node classification task on $\mathcal{G}$. Given labels for a subset of training nodes $\mathcal{D}_{\text{train}} \subset \mathcal{V}$, denoted as $\mathcal{Y}_{\text{train}} = \{\emph{y}_1, \emph{y}_2, \dots, \emph{y}_m\}$, where each $\emph{y}_{i} \in \{1, \dots, |\mathcal{C}|\}$ and $\mathcal{C}$ is the set of all classes. The objective is to learn model parameters $\theta$ such that the trained model $f_\theta$ can accurately predict the labels of nodes in the test set $\mathcal{D}_{\text{test}}$, where $\mathcal{D}_{\text{test}} \cap \mathcal{D}_{\text{train}} = \emptyset$.

\subsection{Graph Unlearning Formalization}
Given an original GNN model $f_{\theta^o}$ trained on a graph $\mathcal{G}$ and an unlearning request $\Delta \mathcal{G}$, the goal of graph unlearning is to obtain an \textit{unlearned} model $f_{\theta^u}$ that closely approximates the \textit{retrained} model $f_{\theta^r}$, where $f_{\theta^r}$ is retrained from scratch on the remaining graph $\mathcal{G}$$\setminus$$\Delta \mathcal{G}$. Graph unlearning can be categorized into three tasks: (1) \textit{Node Unlearning} ($\Delta \mathcal{G}=\{\Delta \mathcal{V}, \emptyset, \emptyset\}$), where a set of nodes (and their associated information) is removed; (2) \textit{Edge Unlearning} ($\Delta \mathcal{G}=\{\emptyset, \Delta \mathcal{E}, \emptyset\}$), where a set of edges is removed; and (3) \textit{Feature Unlearning} ($\Delta \mathcal{G}=\{\emptyset, \emptyset, \Delta \mathcal{X}\}$), where a set of node features is removed, typically masked or replaced with zeros.

\subsection{General Memorization Mechanism}
Memorization was first introduced in computer vision \cite{feldman2020what} to quantify a model’s reliance on specific training samples for prediction. Given a training dataset $\mathcal{D}_{\text{train}}$, training algorithm $\mathcal{T}$, and sample $i \in \mathcal{D}_{\text{train}}$ with feature $\mathbf{x}_{i}$ and label $\emph{y}_i$, the memorization score is:
\begin{equation}
\text{mem}(i) = 
\Pr_{f \sim \mathcal{T}(\mathcal{D}_{\text{train}})} \left[ f(\mathbf{x}_{i}) = \emph{y}_i\right] 
- \Pr_{f \sim \mathcal{T}(\mathcal{D}_{\text{train}} \setminus i)} \left[ f(\mathbf{x}_{i}) = \emph{y}_i\right],
\label{ori_mem}
\end{equation}
where the first term measures the prediction probability with sample $i$ included in training, while the second measures it when $i$ is excluded. A higher score indicates stronger memorization. Although well studied in Euclidean domains, extending this mechanism to graphs requires considering the impact of graph connectivity.

\section{Methodology}
\subsection{Framework Overview}
MGU consists of three main components: (1) \textit{Memorization-based difficulty assessment} leverages memorization strength to estimate unlearning difficulty; (2) \textit{Adaptive unlearning strategy} defines adaptive unlearning objectives based on task difficulty and generalization impact; (3) \textit{Comprehensive evaluation protocol} employs difficulty-aware sampling and task-aware metrics to assess the robustness, practicality, and generalizability of GU methods. 


\subsection{Memorization-based Difficulty Assessment}
\label{4.2}
\noindent \textbf{Memorization Evaluation on Graphs.} 
We first quantify the memorization degree of graph elements in GNNs. The memorization score in Eq. (\ref{ori_mem}) originates from Euclidean domains, where training samples are assumed to be independent. However, this assumption fails in graphs, where a node’s prediction depends not only on its own features but also on information propagated from its neighbors. Therefore, memorization evaluation on graphs must consider both the direct effect on the target node and the indirect influence on its neighbors. To capture these dual effects, we extend Eq. (\ref{ori_mem}) by incorporating a neighbor-impact term:
\begin{equation}
\operatorname{mem}(v_i)
= \alpha\cdot|\Delta(v_i)| + (1-\alpha)\cdot\Delta{\mathcal{N}_k(v_i)},
\label{eq2}
\end{equation}
where $\Delta(v_i)$ is defined following Eq. (\ref{ori_mem}) to measure the change in prediction probability when $v_i$ is excluded from $\mathcal{D}_{\mathrm{train}}$. $\Delta{\mathcal{N}_k(v_i)}$ aggregates the impact on its $k$-hop neighbors ($k$ equals the number of GNN layers). $\alpha\in[0,1]$ balances the two terms, and absolute values capture their magnitudes. Specifically, $\Delta{\mathcal{N}_k(v_i)}$ is computed as the weighted sum of prediction changes across neighbors:
\begin{equation}
\begin{aligned}
\Delta{\mathcal{N}_k(v_i)}
=  \sum\nolimits_{v_\emph{j}\in \mathcal{N}_k(v_i)} &w_{i\emph{j}} \Big(\Big|\Pr_{f\sim \mathcal{T}(\mathcal{D}_{\mathrm{train}})}\left[f(\mathbf{x}_\emph{j})=\emph{y}_\emph{j}\right] \\
& - \Pr_{f\sim \mathcal{T}(\mathcal{D}_{\mathrm{train}}\backslash v_i)}\left[f(\mathbf{x}_\emph{j})=\emph{y}_\emph{j}\right]\Big|\Big),
\end{aligned}
\end{equation}
where the weight $w_{i\emph{j}}$ decays exponentially with graph distance:
\begin{equation}
w_{i\emph{j}}
=\frac{\beta^{\,\operatorname{dis}(v_i,v_\emph{j})}}{\sum\nolimits_{v_t\in \mathcal{N}_k(v_i)} \beta^{\,\operatorname{dis}(v_i,v_t)}},
\end{equation}
where $\operatorname{dis}(v_i,v_\emph{j})$ denotes the shortest path length between $v_i$ and $v_\emph{j}$, and $\beta\in(0,1]$ controls the decay rate over distance. The effects and selection of $\alpha$ and $\beta$ are discussed in Appendix.

\noindent \textbf{Proxy for GU Difficulty.} Next, we define a proxy metric to characterize unlearning difficulty as a unified basis for analysis. The goal is to quantify the difficulty of achieving the \textit{trade-off of unlearning}: forgetting deleted elements ($\Delta \mathcal{G}$), preserving performance on the remaining graph ($\mathcal{G}$$\setminus$$\Delta \mathcal{G}$), and maintaining generalization on the test set $\mathcal{D}_{\text{test}}$. Existing GU studies typically evaluate these objectives separately \cite{li2025toward, yang2024erase}, lacking a unified metric to capture their joint effect. To bridge this gap, inspired by unified evaluation frameworks in related domains \cite{triantafillou2024we}, we propose the \textit{trade-off of unlearning} (ToU) metric, which measures the relative performance difference between the unlearned model $f_{\theta^{u}}$ and the retrained model $f_{\theta^{r}}$ across deleted elements, remaining graph, and the test set:
\begin{align}
\operatorname{ToU}(\theta^{u}, & \, \theta^{r}, \Delta\mathcal{G}, \mathcal{G}\backslash\Delta \mathcal{G}, \mathcal{D}_{\text{test}})
= (1 - \operatorname{diff}(\theta^{u}, \theta^{r}, \Delta\mathcal{G})) \notag \\
& \cdot (1 - \operatorname{diff}(\theta^{u}, \theta^{r}, \mathcal{G}\backslash\Delta \mathcal{G}))  \cdot (1 - \operatorname{diff}(\theta^{u}, \theta^{r}, \mathcal{D}_{\text{test}})),
\label{mem_gnn}
\end{align}
where $\operatorname{diff}(\theta^u, \theta^r, \mathcal{D})=|\operatorname{acc}(\theta^u, \mathcal{D}) - \operatorname{acc}(\theta^r, \mathcal{D})|$ is the absolute accuracy difference between $f_{\theta^{u}}$ and $f_{\theta^{r}}$ on dataset $\mathcal{D}$. In node unlearning, for $\Delta \mathcal{G}$ and $\mathcal{G}$$\setminus$$\Delta \mathcal{G}$, $\mathcal{D}$ corresponds to their respective training samples. Intuitively, ToU favors unlearned models that behave similarly to retrained ones across all three sets. The score lies in $[0, 1]$, with higher values indicating more effective unlearning.

\begin{figure}[]
\centering
\scalebox{0.349}{\includegraphics{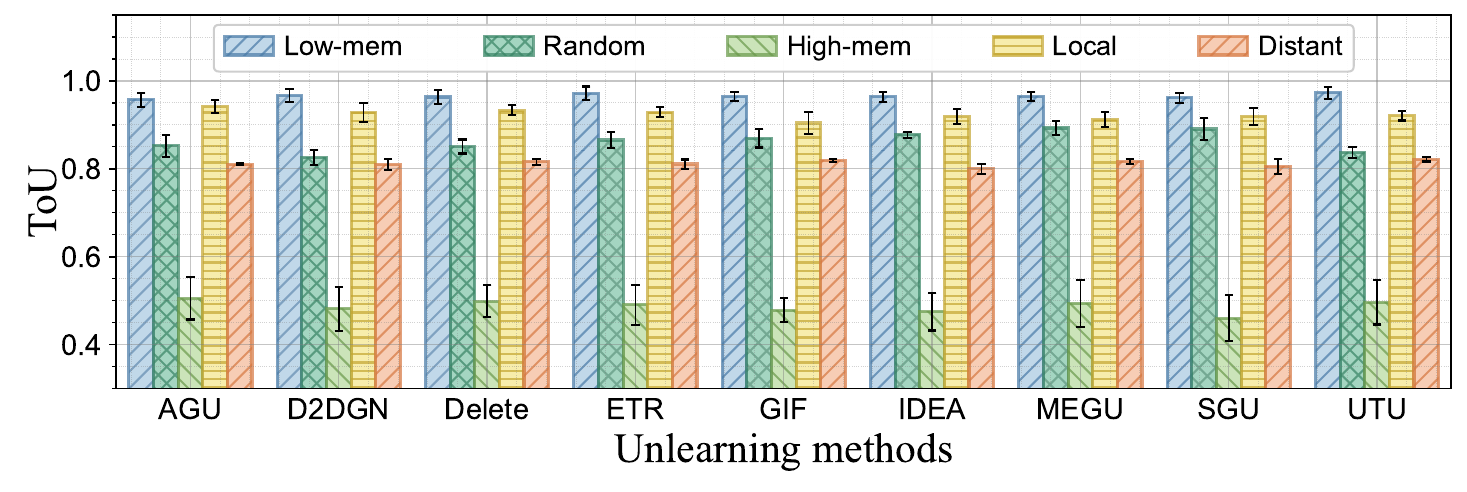}}
\caption{ToU performance of GU methods for node unlearning on Cora with GAT backbone ($N=5$).}
\label{ToU_baseline}
\end{figure}

\noindent \textbf{Relating Memorization to Unlearning Difficulty.} 
We now investigate how GNN memorization correlates with GU difficulty. Our hypothesis is that weakly memorized samples are easier to unlearn, while highly memorized ones are harder. To test this hypothesis against the prevailing assumption that nodes closer to $\mathcal{D}_{\text{test}}$ are more difficult to unlearn, we construct five unlearning sets $\Delta\mathcal{G}$: (1) \textit{low-mem}: the bottom $N\%$ of training nodes with the lowest $\text{mem}(\cdot)$ scores; (2) \textit{high-mem}: the top $N\%$ with the highest scores; (3) \textit{random}: $N\%$ of training nodes randomly sampled, following common practice in existing GU studies; (4) \textit{local}: the $N\%$ of training nodes closest to $\mathcal{D}_{\text{test}}$; and (5) \textit{distant}: the $N\%$ farthest from $\mathcal{D}_{\text{test}}$.

Figure \ref{ToU_baseline} presents three key observations. First, all methods perform best on the \textit{low-mem} set and worst on the \textit{high-mem} set, confirming them as the easiest and hardest to unlearn, respectively. Second, performance on the \textit{random} set is closer to the \textit{low-mem} set than to the \textit{high-mem} set, reflecting the long-tailed nature of memorization difficulty. Third, the \textit{distant} set is harder to unlearn than the \textit{local} set, but both differ significantly from the \textit{high-mem} set, indicating that distance to $\mathcal{D}_{\text{test}}$ does not fundamentally determine unlearning difficulty. These trends are consistently observed across diverse datasets and GNN backbones (see Appendix).

Based on these results, we conclude that the memorization score is a reliable indicator of unlearning difficulty. Therefore, in node unlearning tasks, we define the difficulty of each node $v_i$ as its memorization score, formally, $\mathcal{H}_{\text{node}}(v_i) = \operatorname{mem}(v_i)$.

\noindent \textbf{Quantifying Edge and Feature Unlearning Difficulty.}
We extend the memorization score to edge and feature unlearning. In node classification tasks, edges and features are not directly associated with prediction labels, making the label-based memorization score (Eq. (\ref{mem_gnn})) inapplicable. To overcome this, we infer their unlearning difficulty from the memorization levels of associated nodes. The key insight is that removing edges or features alters the input information that supports label prediction. Thus, when highly memorized nodes heavily rely on the removed elements, preserving predictive performance after unlearning becomes substantially more difficult.

Formally, for each edge $(v_i, v_\emph{j})$, we define its unlearning difficulty by aggregating the memorization levels of its incident nodes (i.e., the two endpoints), normalized by their structural importance:
\begin{equation}
\mathcal{H}_{\mathrm{edge}}(v_i, v_\emph{j}) 
= \frac{\mathcal{H}_{\text{node}}(v_i)}{\sqrt{\operatorname{deg}(v_i)}}
+ \frac{\mathcal{H}_{\text{node}}(v_\emph{j})}{\sqrt{\operatorname{deg}(v_\emph{j})}},
\label{edge_diff}
\end{equation}
where $\operatorname{deg}(v_i)$ denotes the degree of $v_i$. Intuitively, edges connecting highly memorized and low-degree nodes are harder to unlearn, as their removal more severely impacts the model’s predictions on the corresponding nodes. Similarly, for a feature $\mathbf{x}$, let $\mathcal{V}_\mathbf{x}$ be the set of training nodes associated with $\mathbf{x}$. Its unlearning difficulty is defined as the average memorization score of these nodes:
\begin{equation}
\mathcal{H}_{\text{feat}}(\mathbf{x})
= \frac{1}{|\mathcal{V}_\mathbf{x}|}\sum\nolimits_{v_i \in \mathcal{V}_\mathbf{x}}
{\mathcal{H}_{\text{node}}(v_i)}.
\label{feat_diff}
\end{equation}

We validate these metrics through experiments and demonstrate their accuracy in assessing unlearning difficulty (see Appendix).

\subsection{Adaptive Unlearning Strategy}
In this section, we analyze the characteristics of unlearning targets with different difficulty levels and propose an adaptive strategy. As shown in Table \ref{Centrality}, hard-to-unlearn nodes exhibit lower \textit{centrality} values \cite{borgatti2006graph} than easy ones, indicating that they are typically located near the graph boundary. Due to sparse connectivity and weak neighbor aggregation, their features become less representative and harder to generalize. This also explains why nodes farther from the test set (i.e., \textit{distant} nodes) tend to be more difficult to unlearn, as they are usually located in less central regions of the graph.

\begin{figure}[]
\centering
\scalebox{0.345}{\includegraphics{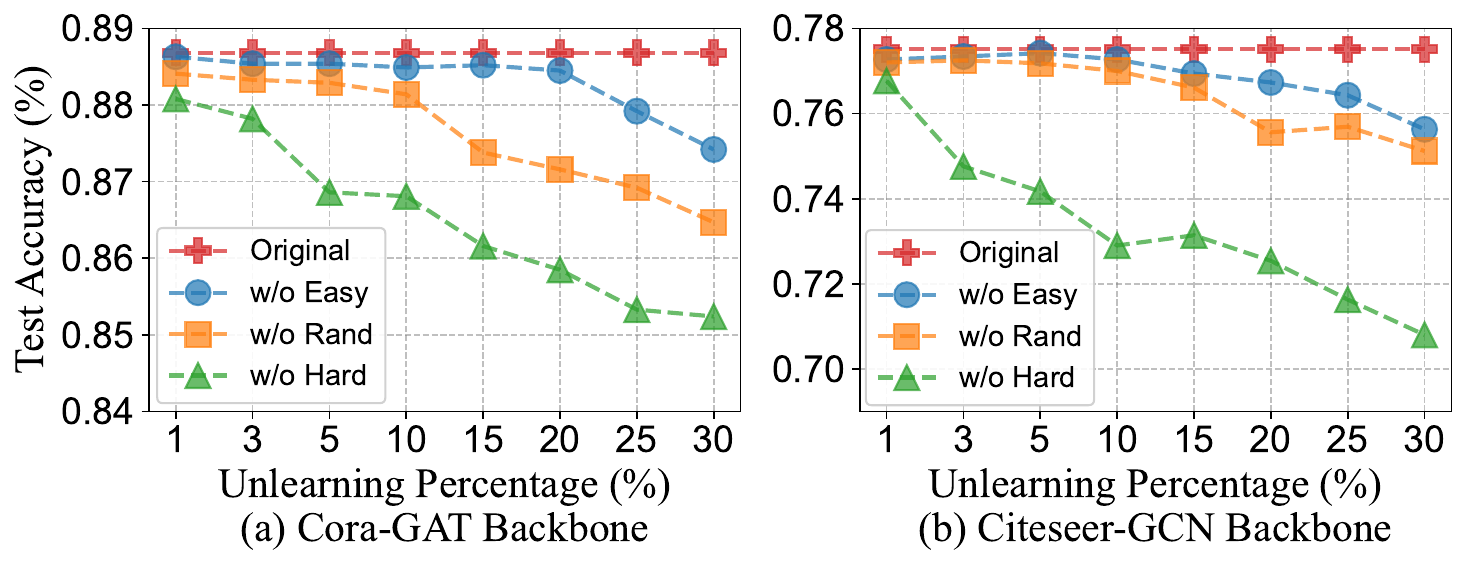}}
\caption{Impact on generalization after removing nodes at varying difficulty levels.}
\label{general_infl}
\end{figure}

Although hard-to-unlearn samples lie near the graph boundary, they are crucial for GNN generalization. As shown in Figure \ref{general_infl}, we retrain models after removing nodes of different difficulty levels and compare their test performance with the original model. Removing hard-to-unlearn nodes leads to substantial degradation, whereas removing the same number of random or easy-to-unlearn nodes has little effect at small unlearning ratios. This shows that unlearning targets at different difficulty levels cause varying generalization loss, challenging the common assumption that the unlearned model should preserve the same generalization as the original one \cite{fan2025opengu}.

\begin{table}[]
\centering
\caption{Centrality values of easy vs. hard-to-unlearn nodes.}
\label{Centrality}
\resizebox{85mm}{!}{
\setlength{\tabcolsep}{1.1mm}{
\begin{tabular}{c|ccc|ccc}
\specialrule{0.05em}{1pt}{1pt} 
\multirow{2}{*}{\begin{tabular}[c]{@{}c@{}}Centrality\\ Metric\end{tabular}} & \multicolumn{3}{c|}{Cora-GAT ($N$=3)} & \multicolumn{3}{c}{Citeseer GCN ($N$=5)} \\
& Easy       & Hard      & Easy/Hard   & Easy       & Hard      & Easy/Hard    \\
\specialrule{0.05em}{1pt}{1pt} 
Degree                                                                       & 1.77e-3   & 9.76e-4   & 1.81       & 2.36e-3     & 3.50e-4    & 6.76        \\
Betweenness                                                                  & 2.15e-3   & 8.55e-4   & 2.52       & 5.05e-3     & 3.90e-6    & 1295.89     \\
Eigenvector                                                                  & 6.61e-3   & 1.38e-3   & 4.79       & 0.02        & 2.42e-5    & 777.62      \\
PageRank                                                                     & 4.42e-4   & 3.09e-4   & 1.43       & 6.01e-4     & 3.16e-4    & 1.9         \\
K-core                                                                       & 2.27      & 1.8       & 1.26       & 2.58        & 1.01       & 2.55       
\\\specialrule{0.05em}{1pt}{1pt} 
\end{tabular}}}
\end{table}

A natural solution is to define different unlearning objectives based on task difficulty. However, computing difficulty scores is costly and requires recomputation after every graph modification. This raises a key question: can we design an efficient approach that dynamically adjusts unlearning objectives based on task difficulty? To this end, we propose an adaptive strategy with two components: (1) margin-based forgetting that adaptively regulates model generalization, and (2) distillation-based preservation that adaptively maintains predictive performance on the remaining graph.

\noindent\textbf{Margin-based Forgetting.}
Prior work has shown that the \textit{margin} between data points and decision boundaries is critical for generalization \cite{liu2016large, song2022tam}. In neural classifiers, the margin is typically defined as the difference between the true class score and other-class scores, where a larger margin implies stronger generalization. Building on this insight, we use margin shifts to capture generalization changes after unlearning. The key idea is that unlearning alters the model’s generalization on deleted elements, with hard-to-unlearn tasks causing greater shifts than random or easy ones.

\begin{figure}[]
\setlength{\abovecaptionskip}{4pt} 
\setlength{\belowcaptionskip}{0pt} 
\centering
\scalebox{0.50}{\includegraphics{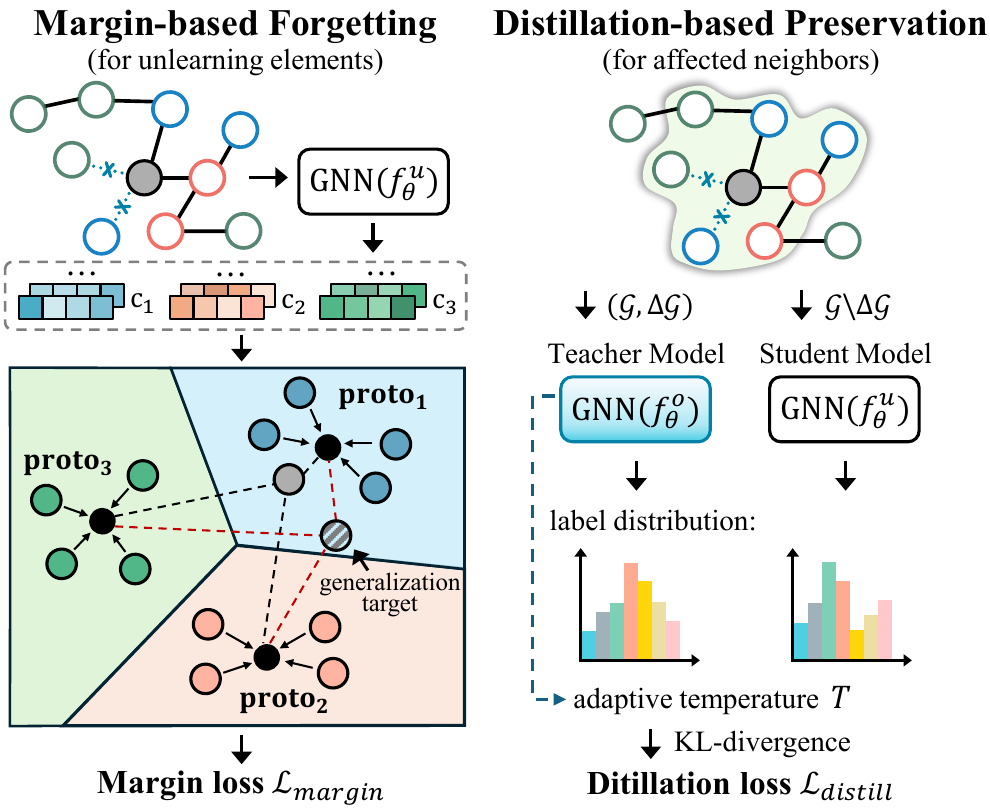}}
\caption{Adaptive Unlearning Strategy.}
\label{model}
\end{figure}

In this work, we propose a prototype-based margin adjustment strategy, where the margin is defined as the difference between a sample’s distance to its true-class prototype and its distance to other-class prototypes. Traditional approaches compute prototypes by averaging embeddings, but in practical unlearning scenarios, the model may be a black box with only probability outputs available. Therefore, we represent each class prototype as the mean probability distribution of its training samples and measure distances using KL divergence. Specifically, let $\mathbf{h}_i$ denote the probability output of node $v_i$. For each class $c$, the prototype is computed as follows:
\begin{equation}
    \mathbf{proto}_c = \frac{1}{|\mathcal{S}_c|} \sum\nolimits_{v_i \in \mathcal{S}_c} \mathbf{h}_i,
\end{equation}
where $\mathcal{S}_c$ is the set of training nodes in class $c$. Then, the margin of $v_i$ is defined as the difference between its distance to the true-class prototype and the average distance to all other-class prototypes:
\begin{equation}
    \gamma_i = \frac{1}{|\mathcal{C}|-1} \sum\nolimits_{c \neq \emph{y}_i} \operatorname{KL}\left(\mathbf{h}_i \,\|\, \mathbf{proto}_c\right)-\operatorname{KL}\left(\mathbf{h}_i  \,\|\,\mathbf{proto}_{\emph{y}_i}\right),
\label{margin_fun}
\end{equation}
where $\mathcal{C}$ is the set of all classes. Since larger margins imply stronger generalization guarantees, $\gamma_i$ serves as a direct indicator of the model's generalization strength. Therefore, we design a margin-based loss as the unlearning objective for the deleted elements:
\begin{equation}
\begin{aligned}
\mathcal{L}_{\text{margin}} 
&= \frac{1}{|\Delta\mathcal{G}|} 
\sum\nolimits_{v_i \in \Delta\mathcal{G}} 
 \max \left(0,\; \gamma_i^{\text{unlearn}} - \left(\gamma_i^{\text{ori}} - \delta_i\right) \right) \\
&\approx \frac{1}{|\Delta\mathcal{G}|}       
\sum\nolimits_{v_i \in \Delta\mathcal{G}} 
 \operatorname{softplus} \left(\gamma_i^{\text{unlearn}} - \tau_i \right),
\end{aligned}
\end{equation}
where $\gamma_i^{\text{ori}}$ and $\gamma_i^{\text{unlearn}}$ are the generalization margins before and after unlearning, $\delta_i$ is the expected margin reduction, and $\tau_i$ is the target margin after unlearning. Since GNN architectures vary widely and trained models are often black-box in practical unlearning scenarios \cite{panda2024fast}, computing a universal $\tau_i$ is intractable. Thus, we treat $\tau_i$ as a learnable parameter initialized from the original model $f_{\theta^o}$ and updated during unlearning. Inspired by prior work \cite{li2024towards, ding2025adaptive}, which shows that the trained model $f_{\theta^o}$ provides self-supervised signals during unlearning, we freeze $f_{\theta^o}$, feed the remaining graph $\mathcal{G}$$\setminus$$\Delta \mathcal{G}$ as input, and apply Eq.~(\ref{margin_fun}) to compute the initial value for each node. We replace non-differentiable $\max(\cdot)$ with its smooth approximation $\operatorname{softplus}(\cdot)$ \cite{zheng2015improving} to ensure stable optimization.

\noindent\textbf{Distillation-based Preservation.} Existing GU methods require the unlearned model $f_{\theta^{u}}$ to preserve the predictive performance of the original model $f_{\theta^{o}}$ on the remaining graph $\mathcal{G}$$\setminus$$\Delta \mathcal{G}$. However, they overlook that deleted elements can indirectly influence predictions on $\mathcal{G}$$\setminus$$\Delta \mathcal{G}$ by altering the model’s generalization. We propose an adaptive distillation strategy that flexibly controls the extent of performance preservation. Here, the original model $f_{\theta^{o}}$ acts as the teacher and the unlearned model $f_{\theta^{u}}$ as the student. Each node is assigned a temperature $T$ reflecting its sensitivity to deletion: highly affected nodes use larger $T$ to smooth the teacher’s distribution and transfer richer inter-class knowledge, while less affected nodes use $T$ close to $1$ for stronger supervision. To estimate node sensitivity, we compare the outputs of $f_{\theta^{o}}$ on the complete graph $\mathcal{G}$ and the remaining graph $\mathcal{G}$$\setminus$$\Delta \mathcal{G}$, and compute $T$ accordingly:
\begin{equation}
T_i = 1 + (T_{\max}-1)\cdot\operatorname{sigmoid}\left(\operatorname{KL}\left(\hat{\mathbf{h}}_i \,\|\, \mathbf{h}_i\right)\right),
\end{equation}
where $\hat{\mathbf{h}}_i$ and $\mathbf{h}_i$ denote the probability distributions of $v_i$ under $f_{\theta^{o}}$ with inputs $\mathcal{G}$$\setminus$$\Delta \mathcal{G}$ and $\mathcal{G}$, respectively. $T_{\max}$ sets the upper bound of the temperature, controlling the softness of the distribution while avoiding over-smoothing. The distillation loss is then defined as:
\begin{equation}
\mathcal{L}_{\text{distill}}
= \frac{1}{|\mathcal{G}\setminus\Delta\mathcal{G}|} \sum_{v_i \in \mathcal{G}\setminus\Delta\mathcal{G}}
\operatorname{KL}\!\left(
\sigma\!\left({\mathbf{z}_i^{u}}/{T_i}\right)
\;\Big\|\;
\sigma\!\left({\mathbf{z}_i^{o}}/{T_i}\right)
\right)\cdot T_i^2,
\end{equation}
where ${\mathbf{z}_i^{u}}$ and ${\mathbf{z}_i^{o}}$ represent the logits of the student (unlearn model) and teacher (original model), respectively, and $\sigma(\cdot)$ denotes the softmax function. The overall training objective is obtained by jointly optimizing the margin and distillation losses:
\begin{equation}
\mathcal{L} = \lambda\cdot\mathcal{L}_{\operatorname{\text{margin}}} + \mathcal{L}_{\operatorname{\text{distill}}},
\end{equation}
where $\lambda$ is a balancing coefficient between the two objectives.

\subsection{Comprehensive Evaluation Framework}
\label{4.4}
In this section, we propose a difficulty-aware sampling strategy and a task-aware evaluation protocol to comprehensively evaluate the robustness, practicality, and generalizability of GU methods.

\noindent\textbf{Difficulty-aware Sampling Strategy.} While our difficulty assessment addresses the limitation of existing evaluations that focus on easy-to-unlearn targets and overlook hard ones, computing these scores still takes time and may not suit real-world scenarios. This motivates an efficient strategy to estimate difficulty scores.

Since our assessment relies on estimating memorization scores $\operatorname{mem}(\cdot)$, a natural approach is to develop fast approximations and extend them to difficulty estimation. Some studies \cite{zhao2024scalability} approximate memorization using proxy indicators such as confidence or accuracy changes during training, but these methods require retraining and are unsuitable for graphs. A recent graph-based approach \cite{jamadandi2025memorization} estimates memorization via node homophily and heterophily, but it requires complete ground-truth labels, which are often unavailable.

To address these limitations, we propose a training-free method that estimates unlearning difficulty directly from the original model $f_{\theta^o}$ and the graph $\mathcal{G}$. Based on our finding that generalization capacity correlates with unlearning difficulty, we adopt the node margin (Eq. (\ref{margin_fun})) as a difficulty indicator, i.e., $\mathcal{H}_{\text{node}}(v_i)\approx\gamma_i$, and extend it to edge and feature difficulties via Eqs. (\ref{edge_diff}) and (\ref{feat_diff}). We validate this approximation through correlation analysis (see Appendix).

\begin{table*}[]
\setlength{\abovecaptionskip}{2pt} 
\setlength{\belowcaptionskip}{0pt} 
\centering
\caption{Node unlearning performance comparison. The best results are in \textbf{bold}, and the second-best results are \underline{underlined}.}
\label{exp_all}
\resizebox{178mm}{!}{
\setlength{\tabcolsep}{0.8mm}{
\begin{tabular}{c|cc|cc|cc|cc|cc|cc|cc|cc|cc|cc}
\specialrule{0.05em}{1pt}{1pt} 
\textbf{Method}  & \multicolumn{2}{c|}{\textbf{Cora}}            & \multicolumn{2}{c|}{\textbf{Citeseer}}        & \multicolumn{2}{c|}{\textbf{PubMed}}          & \multicolumn{2}{c|}{\textbf{Photo}}          & \multicolumn{2}{c|}{\textbf{Computers}}       & \multicolumn{2}{c|}{\textbf{CS}}              & \multicolumn{2}{c|}{\textbf{Physics}}       & \multicolumn{2}{c|}{\textbf{Arxiv}} & \multicolumn{2}{c|}{\textbf{Chameleon}}       & \multicolumn{2}{c}{\textbf{Squirrel}}         \\
\textbf{Bone}    & \textbf{GCN}              & \textbf{GAT}              & \textbf{GCN}              & \textbf{GAT}              & \textbf{GCN}              & \textbf{GAT}              & \textbf{GCN}              & \textbf{GAT}             & \textbf{GCN}              & \textbf{GAT}              & \textbf{GCN}              & \textbf{GAT}              & \textbf{GCN}             & \textbf{GAT}             & \textbf{GCN}         & \textbf{GAT}         & \textbf{SAGE}             & \textbf{FAGCN}            & \textbf{SAGE}              & \textbf{FAGCN}            \\\specialrule{0.05em}{1pt}{1pt} 
\multicolumn{21}{c}{\cellcolor[HTML]{E5E5E5}$\textbf{{Easy to Unlearn}}$}           \\\specialrule{0.05em}{1pt}{1pt} 
AGU     & 98.1±.9          & 95.7±.6          & 96.6±.9          & {\underline {95.6±.3}}   & 97.7±.1          & 98.6±.5          & 98.8±.1          & {\underline {98.3±.7}}    & 92.9±.2          & 97.6±.6          & 99.6±.2          & 99.0±.4          & {\underline {99.6±.4}}    & 99.1±.4          & 98.4±.2          & {\underline {98.7±.1}}    & {\underline {81.5±.0}}    & {{51.7±3}}    & {\underline {78.0±1}}     & 49.5±2          \\
D2DGN   & 97.0±1           & 96.7±.4          & 96.6±.8          & 95.4±.3         & 99.1±.2          & 97.8±.5          & 98.5±.2          & 96.5±.3          & 91.7±.2          & 96.5±.5          & 99.6±.1          & 99.1±.3          & 99.1±.1          & 98.9±.1          & 97.5±.2          & 98.6±.2          & 79.1±1           & 51.5±4          & 72.6±2           & 39.7±1          \\
Delete  & 97.7±.8          & 96.4±.6          & 96.5±1           & 94.5±1          & 98.6±.1          & 98.6±.8          & 96.9±.1          & 96.4±2           & 92.3±.2          & 96.0±.4          & 99.5±.1          & 99.0±.3          & 98.7±.3          & 99.1±.4          & 92.8±1          & 97.9±.3          & 74.1±1           & 47.0±4          & 72.5±.7          & 45.1±1          \\
ETR     & 98.7±.8          & 97.2±.5          & {\underline {96.7±1}}     & 95.2±.5         & 98.5±.2          & 98.2±.3          & {\underline {98.9±.1}}    & 97.4±.3          & 92.9±.2          & 97.8±.7          & {\underline {99.7±.1}}    & {\underline {99.4±.3}}    & 99.5±.2          & 99.1±.2          & 98.1±.5          & 98.4±.3          & 79.1±4           & \underline{54.1±5}          & 76.8±2           & 39.9±3          \\
GIF     & {\underline {98.8±.7}}    & 96.5±.1          & 96.0±2           & 94.9±1          & 99.2±.2          & {\underline {98.7±.8}}    & 98.7±.1          & 97.8±.5          & 92.0±.2          & 97.6±.6          & 99.5±.2          & 99.1±.4          & 99.5±.3          & 99.2±.4          & 97.7±.1          & 96.4±.3               & 79.3±.5          & 45.6±3          & 77.5±1           & 43.4±.9         \\
IDEA    & 88.4±2           & 96.4±.2          & 89.2±4           & 93.8±.7         & 89.4±3           & 96.6±.7          & 97.3±.3          & 98.3±.5          & 92.9±.2          & 96.6±.4          & 99.6±.1          & 99.0±.3          & 99.2±.2          & 98.9±.2          & 88.8±2          & 91.4±.3               & 76.5±.7          & 48.7±4          & 77.6±1           & 44.8±.9         \\
MEGU    & 97.7±.8          & 96.5±1           & 96.5±.5          & 94.4±.9         & 98.7±.1          & 98.6±.5          & 98.8±.1          & 98.1±.6          & {\underline {93.0±.3}}    & 97.5±.5          & 99.5±.1          & 98.8±.3          & 99.5±.4          & 98.9±.1          & {\underline {98.5±.2}}    & 98.2±.2          & 80.8±.5          & 47.9±4          & 73.2±2           & {\underline {51.2±3}}    \\
SGU     & 97.4±.8          & 96.2±1           & 96.4±.9          & 93.6±.6         & {\underline {99.4±.1}}    & 98.3±.3          & 98.8±.1          & 98.1±.6          & 92.9±.2          & {\underline {98.3±.2}}    & 99.5±.1          & 98.9±.2          & 99.4±.0          & {\underline {99.3±.1}}    & 93.0±.5          & 91.5±.4          & 78.8±.4          & 51.5±7          & 73.4±.8          & 43.3±2          \\
UTU     & 97.9±1           & {\underline {97.3±.4}}    & 94.7±.4          & 94.9±1          & 97.6±.2          & 98.6±.6          & 96.4±.1          & 96.2±.5          & 92.9±.2          & 96.5±.4          & 99.5±.1          & 99.0±.2          & 99.5±.4          & 99.2±.4          & 96.4±.2          & 97.3±.2          & 77.9±.9          & 50.8±5          & 72.7±4           & 41.6±4          \\
MGU     & \textbf{98.9±.7} & \textbf{98.4±.8} & \textbf{98.5±.7} & \textbf{96.7±1} & \textbf{99.6±.1} & \textbf{99.5±.2} & \textbf{99.5±.1} & \textbf{98.8±.5} & \textbf{93.5±.7} & \textbf{98.8±.4} & \textbf{99.9±.1} & \textbf{99.6±.2} & \textbf{99.9±.1} & \textbf{99.8±.1} & \textbf{98.7±.5} & \textbf{99.3±.3} & \textbf{84.4±1}  & \textbf{61.0±6} & \textbf{81.4±5}  & \textbf{69.4±3} \\
Improv.$^1$ & 0.1\%            & 1.1\%            & 1.9\%            & 1.2\%           & 0.2\%            & 0.8\%            & 0.6\%            & 0.5\%            & 0.5\%            & 0.5\%            & 0.2\%            & 0.2\%            & 0.3\%            & 0.5\%            & 0.2\%             & 0.6\%             & 3.5\%            & 12.8\%          & 4.4\%            & 35.6\%        
\\\specialrule{0.05em}{1pt}{1pt} 
\multicolumn{21}{c}{\cellcolor[HTML]{E5E5E5}$\textbf{{Random}}$}            
\\\specialrule{0.05em}{1pt}{1pt} 
AGU     & 88.6±.9          & 85.2±2           & 80.5±.4          & 86.0±.6         & \underline{97.5±.8}    & 97.3±1           & \underline{97.5±.2}    & 96.0±.8          & 95.2±.6          & 87.1±.8          & \underline{99.4±.3}    & 98.4±1           & \underline{98.7±.2}    & 98.8±.3          & 97.9±.3    & \underline{98.1±.2}    & 42.6±1           & 47.5±2          & \underline{50.6±3}     & 35.4±4          \\
D2DGN   & 90.8±.7          & 82.6±1           & 85.4±.6          & 86.4±.4         & 96.7±.4          & 96.2±.7          & 95.5±.4          & 97.3±.9          & 95.6±.4          & 88.3±.5          & 99.3±.2          & 97.9±.2          & 97.9±.1          & 97.9±.1          & 96.1±.3          & 96.3±.2          & \underline{43.8±5}     & 49.5±2          & 48.3±2           & 34.9±3          \\
Delete  & 88.6±.9          & 85.1±.6          & 82.1±.9          & 91.1±3          & 96.4±.7          & 97.9±.6          & 96.5±.3          & 96.9±2           & 95.1±.9          & \underline{89.6±1}     & 97.8±.2          & 98.7±.4          & 97.0±.4          & 98.5±.4          & 87.7±2          & 97.2±.1          & 38.4±1           & 47.0±3          & 45.0±2           & 34.5±2          \\
ETR     & {92.7±.4}    & 86.6±1           & 83.9±1           & \underline{93.6±2}    & 94.5±.4          & \underline{98.6±.4}    & 97.3±.4          & \underline{97.4±1}     & 96.2±.3          & 88.1±1           & 98.7±.2          & \underline{98.8±.5}    & 98.4±.3          & \underline{98.9±.6}    & 97.7±.5          & 97.4±.6          & 36.9±2           & \underline{57.7±1}    & 44.2±1           & \underline{36.4±3}    \\
GIF     & {94.3±.6}          & 86.9±2           & 86.5±1           & 90.4±.3         & 96.3±.2          & 96.5±.8          & 97.3±.3          & 96.1±1           & 96.0±.4          & 87.5±.6          & 98.4±.3          & 98.6±.6          & 97.2±.2          & 98.7±.3          & 96.2±.4          & 94.7±.8               & 36.4±.9          & 51.3±2          & 49.2±2           & 36.0±3          \\
IDEA    & 88.7±.5          & 87.7±.7          & 81.0±.9          & 85.1±2          & 88.3±4           & 96.9±1           & 97.0±.3          & 95.6±.9          & 95.7±.6          & 87.5±.7          & 98.4±.2          & 98.7±.5          & 97.3±.3          & 98.1±.4          & 84.2±5          & 82.7±.9               & 42.6±.5          & 49.8±5          & 47.3±1           & 33.0±6          \\
MEGU    & 91.0±.7          & \underline{89.3±.6}    & \underline{86.8±.9}    & 87.1±3          & 97.2±.5          & 97.0±.9          & 95.3±.4          & 96.3±.9          & \underline{96.6±.5}    & 87.5±.7          & 99.3±.2          & 97.9±.7          & 98.4±.4          & 98.2±.6          & \underline{98.1±.3}          & 97.0±.4          & 39.6±.8          & 51.7±3          & 42.8±1           & 31.6±2          \\
SGU     & \underline{94.7±1}          & 89.1±.5          & 84.0±.7          & 91.8±3          & 97.2±.4          & 96.4±.5          & 97.4±.2          & 96.0±.5          & 96.2±.2          & 88.0±1           & 99.1±.1          & 98.4±.7          & 98.5±.1          & 98.5±.4          & 88.1±.6          & 84.1±.7          & 40.0±2           & 50.5±2          & 40.2±1           & 35.5±1          \\
UTU     & 92.7±.9          & 83.8±1           & 81.0±.6          & 87.8±.8         & 96.1±.4          & 97.1±.7          & 96.6±.4          & 95.8±.9          & 95.9±.4          & 86.1±.9          & 98.6±.2          & 98.2±.6          & 98.2±.2          & 98.6±.9          & 94.8±.4          & 96.0±.3          & 37.5±1           & 51.6±2          & 45.3±2           & 34.0±2          \\
MGU     & \textbf{97.8±.3} & \textbf{96.0±.4} & \textbf{95.7±1}  & \textbf{95.7±1} & \textbf{98.5±.2} & \textbf{98.8±.4} & \textbf{98.7±.4} & \textbf{98.6±.7} & \textbf{97.0±1}  & \textbf{98.5±.3} & \textbf{99.5±.1} & \textbf{99.2±.2} & \textbf{99.2±.1} & \textbf{99.6±.2} & \textbf{98.9±.5} & \textbf{99.1±.2} & \textbf{60.2±.6} & \textbf{61.7±3} & \textbf{69.2±.8} & \textbf{63.7±1} \\
Improv. & 3.3\%            & 7.5\%            & 10.3\%           & 2.2\%           & 1.0\%            & 0.2\%            & 1.2\%            & 1.2\%            & 0.4\%            & 9.9\%            & 0.1\%            & 0.4\%            & 0.5\%            & 0.7\%            & 0.82\%            & 1.02\%            & 37.4\%           & 6.9\%           & 36.8\%           & 75.1\%        
\\\specialrule{0.05em}{1pt}{1pt} 
\multicolumn{21}{c}{\cellcolor[HTML]{E5E5E5}$\textbf{{Hard to Unlearn}}$}  
\\\specialrule{0.05em}{1pt}{1pt} 
AGU     & 43.4±3           & \underline{50.5±4}     & 43.9±.5          & 52.8±2          & 80.2±2           & 84.3±2           & 63.4±.2          & 80.4±1           & \underline{63.6±.8}    & 58.1±1           & 81.4±1           & 86.4±2           & 91.0±2           & 89.9±3           & 89.3±.1          & 89.5±.3          & 25.8±1           & 36.0±2          & \underline{20.1±3}     & 26.4±1          \\
D2DGN   & 43.0±2           & 48.1±5           & 44.7±.8          & 52.0±2          & 76.1±2           & 84.8±3           & 63.1±.3          & 81.4±2           & 62.7±.6          & \underline{61.3±1}     & 82.0±.8          & \underline{88.1±1}     & 88.2±.5          & 90.2±.5          & 88.9±.6          & 89.5±.2          & \underline{33.2±1}     & 41.4±4          & 19.7±1           & 30.3±.3         \\
Delete  & 43.1±4           & 49.9±3           & 43.7±.5          & 52.8±2          & 78.6±.6          & 86.1±.6          & \underline{64.3±.9}    & \underline{88.5±3}     & 62.3±.4          & 60.9±3           & 81.4±.7          & 87.9±.8          & 89.8±2           & 91.5±1           & 80.8±2         & \underline{91.6±1}    & 23.1±2           & 38.7±2          & 20.0±.9          & 27.3±2          \\
ETR     & 44.8±1           & 49.0±4           & 43.2±1           & 51.3±1          & 77.5±3           & 84.7±3           & 63.9±.5          & 84.1±3           & 63.1±1           & 59.7±2           & \underline{82.4±1}     & 87.5±.9          & 90.8±2           & 90.6±2           & 88.4±.8          & 89.3±.5          & 24.8±.7          & 38.3±5          & 18.5±.3          & 27.6±1          \\
GIF     & 47.7±2           & 47.8±2           & 46.6±1           & \underline{54.0±.5}   & \underline{88.4±2}     & 85.1±.9          & 63.4±.5          & 81.3±2           & 62.9±.8          & 56.7±1           & 81.1±.7          & 85.7±2           & \underline{91.6±.4}    & 90.1±2           & 88.8±.3          & 83.3±.1               & 25.9±.7          & 38.7±3          & 18.6±1           & 26.5±4          \\
IDEA    & \underline{65.8±5}     & 47.4±4           & \underline{61.6±3}     & 52.5±2          & 87.1±4           & 83.9±2           & 63.4±.4          & 82.4±2           & 63.2±.7          & 57.4±.9          & 81.7±.9          & 87.6±1           & 90.8±.6          & 90.1±3           & \underline{90.7±2}    & 79.7±1               & 24.9±2           & 36.6±4          & 18.3±.3          & 29.2±7          \\
MEGU    & 43.0±3           & 49.4±5           & 43.3±1           & 49.5±1          & 78.7±2.          & 84.8±3           & 63.4±.4          & 81.3±2           & 63.3±.6          & 58.1±2           & 81.6±1           & 85.7±3           & 90.3±2           & 91.6±3           & 89.6±.1          & 90.3±.2          & 24.6±2           & 34.7±2          & 17.8±.7          & 25.1±1          \\
SGU     & 40.8±2           & 46.0±5           & 43.3±.4          & 49.6±2          & 73.8±.9          & \underline{86.6±2}     & 63.4±.4          & 80.8±2           & 62.5±.9          & 58.3±2           & 80.8±.6          & 85.4±2           & 91.1±.3          & \underline{94.0±1}     & 83.0±1          & 85.6±1          & 25.6±1           & \underline{42.2±2}    & 17.5±.4          & \underline{32.4±2}    \\
UTU     & 43.2±2           & 49.7±5           & 43.0±.4          & 51.1±.9         & 78.5±3           & 85.6±2           & 63.2±.2          & 79.8±1           & 63.2±.5          & 57.4±.9          & 81.8±1           & 86.3±2           & 90.4±2           & 89.7±2           & 87.7±.1          & 90.5±.1          & 25.1±.7          & 36.2±2          & 17.8±.5          & 26.6±2          \\
MGU     & \textbf{97.2±1}  & \textbf{95.2±2}  & \textbf{94.8±.6} & \textbf{93.8±2} & \textbf{97.4±.9} & \textbf{97.7±1}  & \textbf{90.6±.9} & \textbf{96.8±1}  & \textbf{96.2±1}  & \textbf{95.8±2}  & \textbf{96.2±.5} & \textbf{98.9±.9} & \textbf{99.6±.2} & \textbf{98.5±1}  & \textbf{97.1±.3} & \textbf{97.5±.1} & \textbf{65.5±2}  & \textbf{69.4±1} & \textbf{70.1±3}  & \textbf{63.8±5} \\
Improv. & 47.7\%           & 88.5\%           & 53.9\%           & 73.7\%          & 10.2\%           & 12.8\%           & 40.9\%           & 9.4\%            & 51.3\%           & 56.3\%           & 16.7\%           & 12.3\%           & 8.7\%            & 4.8\%            & 7.1\%             & 6.4\%             & 152\%            & 64.4\%          & 249\%            & 96.9\%        
\\\specialrule{0.05em}{1pt}{1pt} 
\end{tabular}
}}
\scriptsize{\leftline{$^1$ Improvements of MGU over the best-performing baseline.}}
\end{table*}

\noindent\textbf{Task-aware Evaluation Protocol.}
In Section \ref{4.2}, we propose the ToU metric (Eq. (\ref{mem_gnn})) to jointly evaluate (1) forgetting ability, (2) predictive preservation, and (3) generalization capacity in node unlearning. We now extend ToU to edge and feature unlearning tasks. In node classification, these tasks do not remove node labels from $\mathcal{D}_\text{train}$. Instead, the goal of the unlearned model $f_{\theta^u}$ is to forget predictive patterns that previously depended on deleted edges $\Delta\mathcal{E}$ or features $\Delta\mathcal{X}$. Therefore, applying the same evaluation criteria as in node unlearning is inappropriate, and task-specific protocols are required. Specifically, the trade-off of edge unlearning is defined as: 
\begin{align}
\operatorname{ToU}_\text{edge}(f_{\theta^u})
= &(1 - \operatorname{diff}_\text{MIA}(\theta^{u}, \theta^{r}, \Delta\mathcal{E})) 
 \cdot (1 - \operatorname{diff}(\theta^{u}, \theta^{r}, \mathcal{D}_{\text{train}})) \notag \\ & \cdot (1 - \operatorname{diff}(\theta^{u}, \theta^{r}, \mathcal{D}_{\text{test}})),
\label{tou_edge}
\end{align}
where the forgetting ability $\operatorname{diff}_\text{MIA}(\theta^{u}, \theta^{r}, \Delta\mathcal{E})$ is evaluated using membership inference attacks \cite{hu2022membership}, which compare the capability of $f_{\theta^u}$ and $f_{\theta^r}$ to distinguish edges in $\Delta\mathcal{E}$. Predictive preservation is assessed by comparing the predictions of $f_{\theta^u}$ and $f_{\theta^r}$ on all training samples in $\mathcal{D}_{\text{train}}$, differing from existing methods \cite{li2024towards} that assume the inference capability of nodes linked to $\Delta\mathcal{E}$ should be removed. Generalization capacity is measured by model performance on $\mathcal{D}_\text{test}$. Similarly, the trade-off of feature unlearning is defined as:
\begin{align}
\operatorname{ToU}_\text{feat}(f_{\theta^u})
= &(1 - \operatorname{diff}(\theta^{u}, \theta^{r}, \Delta\mathcal{X})) 
 \cdot (1 - \operatorname{diff}(\theta^{u}, \theta^{r}, \mathcal{D}_{\text{train}})) \notag \\ & \cdot (1 - \operatorname{diff}(\theta^{u}, \theta^{r}, \mathcal{D}_{\text{test}})),
\label{tou_feat}
\end{align}
where the forgetting ability is measured by comparing the predictions of $f_{\theta^u}$ and $f_{\theta^r}$ on $\Delta\mathcal{X}$ with and without graph structure:
\begin{equation}
\operatorname{diff}(\theta^u, \theta^r, \Delta\mathcal{X})
= \frac{1}{|\mathcal{I}|} \sum\nolimits_{i \in \mathcal{I}}
\Big| \operatorname{acc}\!\left(f_{\theta^u}(i)\right) 
   - \operatorname{acc}\!\left(f_{\theta^r}(i)\right) \Big|,
\end{equation}
where $\mathcal{I}=\{(\mathcal{G},\Delta\mathcal{X}), (\varnothing,\Delta\mathcal{X})\}$ denotes the two input settings (with and without graph structure). Predictive preservation is measured by comparing predictions on $\mathcal{D}_{\text{train}}$ using the remaining graph $\mathcal{G}$$\setminus$$\Delta \mathcal{X}$, while generalization capacity is evaluated on $\mathcal{D}_\text{test}$.

\section{Experiments}
We conduct extensive experiments to answer the following research questions: \textbf{RQ1:} How does MGU perform compared to state-of-the-art methods across unlearning tasks with different difficulty levels? \textbf{RQ2:} How does each proposed module in MGU contribute to the overall performance? \textbf{RQ3:} Can our adaptive unlearning strategy enhance existing methods when handling hard-to-unlearn tasks? \textbf{RQ4:} How does MGU perform under different unlearning scales and parameter settings?

\subsection{Experimental Setup}
\textbf{Datasets.} We evaluate MGU on ten widely used benchmark datasets: (1) eight homophily graphs: Cora, Citeseer, PubMed \cite{yang2016revisiting}, Photo, Computers, CS, Physics \cite{shchur2018pitfalls}, and ogbn-arxiv \cite{hu2020open}; (2) two heterophily graphs: Chameleon and Squirrel \cite{zenggraphsaint}. Following recent GU studies \cite{cheng2023gnndelete,li2024towards}, we adopt a standard data split with $80\%$ of nodes for training and $20\%$ for testing. Detailed statistics and descriptions of these datasets are provided in Appendix.

\noindent\textbf{Backbone GNNs and Baselines.} We evaluate the adaptability of MGU on five commonly used backbone GNNs: GCN \cite{kipf2016semi}, GAT \cite{velickovic2017graph}, SAGE \cite{hamilton2017inductive}, GIN \cite{xu2018powerful}, and FAGCN \cite{bo2021beyond}, where FAGCN is specifically applied to heterophily graphs. We compare MGU with nine state-of-the-art baselines: AGU \cite{ding2025adaptive}, D2DGN \cite{sinha2023distill}, Delete \cite{cheng2023gnndelete}, ETR \cite{yang2024erase}, GIF \cite{wu2023gif}, IDEA \cite{dong2024idea}, MEGU \cite{li2024towards}, SGU \cite{li2025toward}, and UTU \cite{tan2024unlink}. Detailed descriptions of the GNNs and baselines are provided in Appendix. For all methods, we fix the embedding dimension at $64$ and use $2$ GNN layers. Baseline parameters are initialized with the settings reported in the original papers and further fine-tuned for optimal performance. To ensure a fair comparison, each experiment is repeated ten times and the average results are reported.

\noindent\textbf{Unlearning Tasks.} We evaluate all methods under three difficulty settings (\textit{easy}, \textit{random}, and \textit{hard}). The \textit{easy} and \textit{hard} settings select the bottom- and top-ranked elements by memorization scores, respectively, while the \textit{random} setting uniformly samples elements following the common protocol in existing work \cite{fan2025opengu}. Each setting is applied to three types of GU tasks: (1) \textit{Node unlearning}, where $5\%$ of nodes and their connected edges are deleted; (2) \textit{Edge unlearning}, where $5\%$ of edges are deleted; and (3) \textit{Feature unlearning}, where the features of $5\%$ of nodes are replaced with zeros. This yields nine evaluation scenarios (three difficulty settings $\times$ three tasks). Results for additional unlearning ratios are provided in Appendix.

\noindent\textbf{Evaluation Metrics.} We adopt the proposed \textit{Trade-off of Unlearning} (ToU) as the primary evaluation metric. In addition, we evaluate whether GU methods effectively prevent information leakage by applying two commonly used attack strategies: membership inference attack \cite{zhang2022inference} and poisoning edge attack \cite{wu2023gif}, at node and edge levels. Detailed attack configurations are provided in Appendix.

\begin{table}[]
\setlength{\abovecaptionskip}{2pt} 
\setlength{\belowcaptionskip}{0pt} 
\centering
\caption{Edge and feature unlearning results on hard tasks.}
\resizebox{85mm}{!}{
\setlength{\tabcolsep}{0.8mm}{
\begin{tabular}{c|cc|cc|cc|cc}
\specialrule{0.05em}{1pt}{1pt}
\textbf{Method} & \multicolumn{2}{c|}{\textbf{Citeseer}} & \multicolumn{2}{c|}{\textbf{Photo}}  & \multicolumn{2}{c|}{\textbf{Computers}} & \multicolumn{2}{c}{\textbf{Chameleon}} \\
\textbf{Bone}   & \textbf{SAGE}     & \textbf{GIN}      & \textbf{SAGE}    & \textbf{GIN}     & \textbf{SAGE}      & \textbf{GIN}      & \textbf{SAGE}      & \textbf{GIN}      \\
\specialrule{0.05em}{1pt}{1pt}
\multicolumn{9}{c}{\cellcolor[HTML]{E5E5E5}$\textbf{{Edge Unlearning}}$}  
\\\specialrule{0.05em}{1pt}{1pt}
AGU             & 27.8±.1           & 77.5±.9           & 81.5±.9          & 46.3±2           & 82.0±.6           & 69.3±.0           & 29.6±2             & 62.3±.9           \\
D2DGN           & 27.9±.1           & \underline{82.1±.3}     & 81.2±.3          & \underline{53.6±3}     & \underline{84.4±1}       & 76.7±.6           & \underline{33.8±3}       & \underline{69.8±1}      \\
Delete          & 28.8±.8           & 77.6±.5           & 81.1±.7          & 36.3±5           & 81.3±.4            & 26.8±3            & 27.8±2             & 60.4±2            \\
ETR             & 27.7±.1           & 75.6±.9           & 81.7±.6          & 16.5±4           & 82.4±.9            & 47.9±3            & 30.2±.9            & 59.4±7            \\
GIF             & 34.6±3            & 78.0±.4           & 81.4±.8          & 38.6±1           & 81.7±.7            & 61.9±6            & 29.3±2             & 57.1±3            \\
IDEA            & 35.6±5            & 81.8±1            & 81.2±.6          & 29.4±3           & 81.6±.5            & 47.9±6            & 30.1±1             & 55.3±4            \\
MEGU            & 27.7±.2           & 76.6±.7           & \underline{82.0±1}     & 37.7±7           & 81.1±.4            & 63.0±4            & 29.2±.8            & 61.7±4            \\
SGU             & \underline{42.2±7}      & 78.4±.8           & 80.7±.2          & 46.2±5           & 81.2±.4            & \underline{78.6±.9}     & 30.0±.9            & 68.4±1            \\
UTU             & 27.8±.2           & 77.5±2            & 81.8±.8          & 33.4±2           & 81.9±1             & 52.7±6            & 29.8±.4            & 62.8±2            \\
MGU             & \textbf{94.5±4}   & \textbf{90.8±1}   & \textbf{93.2±.6} & \textbf{83.6±2}  & \textbf{93.9±2}    & \textbf{80.8±.2}  & \textbf{45.8±.3}   & \textbf{71.9±.6}  \\
Improv.         & 124\%             & 10.6\%            & 13.7\%           & 56.0\%           & 11.3\%             & 2.8\%             & 35.5\%             & 3.0\%             \\
\specialrule{0.05em}{1pt}{1pt}
\multicolumn{9}{c}{\cellcolor[HTML]{E5E5E5}$\textbf{{Feature Unlearning}}$}  \\\specialrule{0.05em}{1pt}{1pt}
AGU             & \underline{33.6±.4}     & 77.7±3            & 62.0±2           & 4.3±.6           & 60.0±1             & 47.1±5            & 12.1±.6            & 59.4±1            \\
D2DGN           & 23.4±2            & 76.6±4            & 60.9±1           & 63.1±9           & 59.9±.6            & 54.3±3            & \underline{28.4±2}       & \underline{80.1±7}      \\
Delete          & 2.9±.0            & \underline{84.8±.5}     & 61.5±2           & 7.3±2            & 59.6±.5            & 14.6±.3           & 11.8±.8            & 72.5±6            \\
ETR             & 18.2±6            & 82.7±.7           & 60.8±2           & 25.3±2           & 60.0±.9            & 42.9±5            & 11.8±.1            & 57.6±4            \\
GIF             & 14.1±7            & 76.1±.8           & 61.9±2           & 65.5±2           & 59.1±1             & 61.6±7            & 11.8±.4            & 59.0±4            \\
IDEA            & 2.9±.0            & 74.8±2            & \underline{63.8±4}     & 68.4±5           & \underline{60.2±.6}      & 52.2±3            & 11.7±.2            & 51.1±3            \\
MEGU            & 2.9±.0            & 78.4±3            & 62.8±4           & \underline{70.9±3}     & 59.6±.7            & \underline{64.8±2}      & 11.5±.3            & 64.3±8            \\
SGU             & 18.7±4            & 75.6±1            & 59.0±.6          & 53.1±7           & 57.6±.7            & 62.6±4            & 11.5±.4            & 73.7±7            \\
UTU             & 23.7±5            & 81.5±.4           & 62.3±.7          & 60.8±4           & 59.7±.8            & 63.4±2            & 11.7±.3            & 77.8±4            \\
MGU             & \textbf{87.4±3}   & \textbf{91.1±1}   & \textbf{80.8±.7} & \textbf{89.0±.9} & \textbf{74.0±.5}   & \textbf{73.1±1}   & \textbf{70.1±2}    & \textbf{82.2±2}   \\
Improv.         & 160\%             & 7.4\%             & 26.6\%           & 25.5\%           & 22.9\%             & 12.8\%            & 146\%              & 2.6\%       \\\specialrule{0.05em}{1pt}{1pt}    
\end{tabular}}}
\label{exp_edge_feat}
\end{table}

\subsection{Performance Comparison (RQ1)}
We compare the performance of all methods from three perspectives: \textit{effectiveness}, \textit{efficiency}, and \textit{adversarial robustness}.

\begin{table}[]
\setlength{\abovecaptionskip}{2pt} 
\setlength{\belowcaptionskip}{0pt} 
\centering
\caption{Average runtime of node unlearning (in seconds).}
\resizebox{85mm}{!}{
\setlength{\tabcolsep}{1.2mm}{
\begin{tabular}{c|cccccccc}
\specialrule{0.05em}{1pt}{1pt}
\textbf{Method}  & \textbf{Cora}          & \textbf{Citeseer}      & \textbf{PubMed}        & \textbf{CS}            & \textbf{Physics}       & \textbf{Arxiv}                     & \textbf{Cham.}    & \textbf{Squi.}      \\\specialrule{0.05em}{1pt}{1pt}
\multicolumn{9}{c}{\cellcolor[HTML]{E5E5E5}$\textbf{{GCN}}$}                 \\\specialrule{0.05em}{1pt}{1pt}
Retrain & 0.71          & 0.53          & 0.96          & 2.06          & 2.93          & 16.18         & 0.52          & 0.88          \\\specialrule{0.05em}{1pt}{1pt}
ETR     & \textbf{0.02} & {\ul 0.02}    & \textbf{0.02} & \textbf{0.02} & \textbf{0.03} & \textbf{0.03} & \textbf{0.02} & \textbf{0.01} \\
GIF     & 0.14          & 0.11          & 0.11          & 0.32          & 0.73          & 2.08          & 0.10          & 0.27          \\
IDEA    & 0.10          & 0.11          & 0.11          & 0.32          & 0.73          & 2.13          & 0.10          & 0.21          \\\specialrule{0.05em}{1pt}{1pt}
AGU     & 0.06          & 0.06          & 0.06          & 0.12          & 0.24          & 0.48          & 0.07          & 0.09          \\
D2DGN   & 0.16          & 0.18          & 0.17          & 0.26          & 0.51          & 1.14          & 0.16          & 0.20          \\
Delete  & 1.37          & 1.23          & 1.31          & 2.47          & 4.51          & 39.25         & 1.51          & 3.01          \\
MEGU    & 0.09          & 0.08          & 0.08          & 0.14          & 0.26          & 0.51          & 0.09          & 0.12          \\
SGU     & 0.47          & 0.55          & 2.79          & 2.60          & 5.44          & 28.34         & 0.42          & 0.83          \\
MGU     & {\ul 0.02}    & \textbf{0.01} & {\ul 0.03}    & {\ul 0.07}    & {\ul 0.04}    & {\ul 0.20}    & {\ul 0.04}    & {\ul 0.05}    \\\specialrule{0.05em}{1pt}{1pt}
\multicolumn{9}{c}{\cellcolor[HTML]{E5E5E5}$\textbf{{GAT}}$}                 \\\specialrule{0.05em}{1pt}{1pt}
Retrain & 0.77          & 0.55          & 0.82          & 2.88          & 4.02          & 30.04         & 0.63          & 1.84          \\\specialrule{0.05em}{1pt}{1pt}
ETR     & {\ul 0.03}    & {\ul 0.02}    & \textbf{0.02} & \textbf{0.03} & \textbf{0.05} & \textbf{0.04} & {\ul 0.02}    & \textbf{0.02} \\
GIF     & 0.35          & 0.32          & 0.35          & 0.62          & 1.39          & 82.27         & 0.31          & 3.46          \\
IDEA    & 0.36          & 0.34          & 0.35          & 0.63          & 1.44          & 86.38         & 0.32          & 3.32          \\\specialrule{0.05em}{1pt}{1pt}
AGU     & 0.09          & 0.08          & 0.10          & 0.21          & 0.43          & 1.05          & 0.10          & 0.13          \\
D2DGN   & 0.22          & 0.24          & 0.24          & 0.46          & 0.84          & 1.85          & 0.24          & 0.30          \\
Delete  & 1.85          & 1.82          & 2.03          & 4.05          & 7.79          & 13.17         & 1.90          & 1.95          \\
MEGU    & 0.11          & 0.11          & 0.13          & 0.23          & 0.45          & 1.03          & 0.13          & 0.17          \\
SGU     & 0.50          & 0.58          & 2.89          & 2.74          & 6.00          & 25.43         & 0.46          & 0.84          \\
MGU     & \textbf{0.02} & \textbf{0.02} & {\ul 0.04}    & {\ul 0.11}    & {\ul 0.23}    & {\ul 0.15}    & \textbf{0.02} & \underline{0.03}
\\\specialrule{0.05em}{1pt}{1pt}
\end{tabular}}}
\label{time}
\end{table}

\noindent\textbf{Effectiveness.}
Tables \ref{exp_all} and \ref{exp_edge_feat} report the ToU performance across GU tasks of different difficulty levels and backbone GNNs. MGU consistently outperforms all baselines across all settings. On average, it achieves improvements of 3.29\% on easy tasks (0.1\% to 35.6\%), 9.82\% on random tasks (0.1\% to 75.1\%), and 53.1\% on hard tasks (2.6\% to 249\%) over the best-performing baselines. These results demonstrate MGU's robustness across diverse difficulty levels, task types, backbone GNNs, and graph types (homophily and heterophily), highlighting the advantage of its adaptive unlearning strategy. In contrast, baseline methods adopt uniform unlearning schemes, leading to inferior performance, particularly on hard tasks. 

\begin{figure}[]
\centering
\scalebox{0.35}{\includegraphics{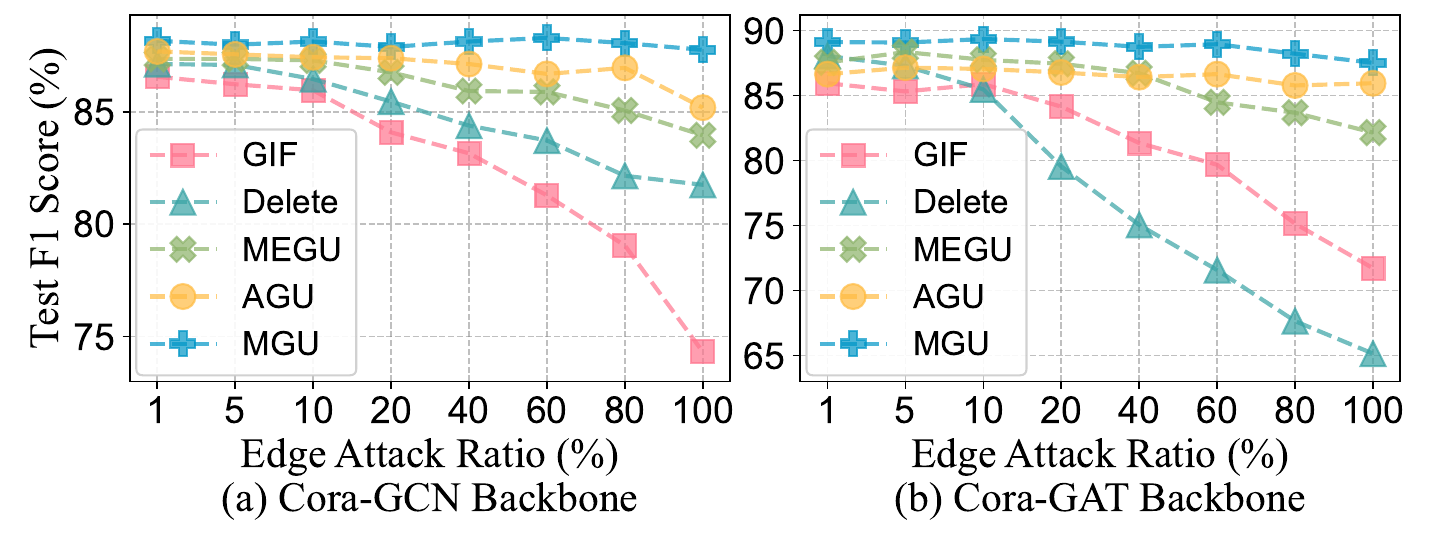}}
\caption{PEA performance under edge unlearning.}
\label{pea}
\end{figure}

\noindent\textbf{Efficiency.}
Table \ref{time} reports the average unlearning time. MGU significantly outperforms learning-based baselines (AGU, D2DGN, Delete, MEGU, and SGU), achieving speedups of $3.5\times$ to $70\times$. Compared with training-free methods (GIF, IDEA, ETR), which directly adjust model parameters without training, MGU remains highly competitive, ranking second only to ETR in some cases. However, ETR requires additional preprocessing time to extract subgraphs based on unlearning requests. Moreover, our experiments show that training-free methods are sensitive to hyperparameters and may cause model collapse on the test set. In contrast, MGU maintains robust and stable performance with only 10–30 training epochs.

\noindent\textbf{Adversarial Robustness.}
Figure \ref{pea} shows the robustness of GU methods against poisoning edge attacks (PEA). Across both GCN and GAT backbones, MGU exhibits superior resistance to adversarial perturbations, maintaining stable performance as the attack ratio increases. In contrast, GIF and Delete experience substantial degradation, with Delete showing particularly sharp drops on GAT. Results for membership inference attacks (MIA) are provided in Appendix, where MGU consistently outperforms all baselines.

\begin{table}[]
\setlength{\abovecaptionskip}{2pt} 
\setlength{\belowcaptionskip}{0pt} 
\centering
\caption{Node unlearning performance of MGU variants.}
\resizebox{85mm}{!}{
\setlength{\tabcolsep}{0.7mm}{
\begin{tabular}{cc|ccc|ccc}
\specialrule{0.05em}{1pt}{1pt}
\multirow{2}{*}{\begin{tabular}[c]{@{}c@{}}\textbf{GNN}\\ \textbf{Bone}\end{tabular}} & \multirow{2}{*}{\textbf{Method}} & \multicolumn{3}{c|}{\textbf{PubMed}}                             & \multicolumn{3}{c}{\textbf{Photo}}                              \\
&                         & \textbf{Easy}             & \textbf{Random}             & \textbf{Hard}             & \textbf{Easy}             & \textbf{Random}             & \textbf{Hard}             \\\specialrule{0.05em}{1pt}{1pt}
\multirow{3}{*}{GCN}                                                & w/o Margin               & 96.5±.8          & 95.5±.9          & 82.9±.6          & 98.7±.3          & 97.5±.2          & 77.4±.4          \\
& w/o Distill             & 97.1±.5          & 94.6±.3          & 83.6±.7          & 95.6±1           & 87.3±8           & 76.3±1           \\
& MGU                     & \textbf{99.6±.1} & \textbf{98.7±.4} & \textbf{90.6±.9} & \textbf{99.5±.1} & \textbf{98.7±.4} & \textbf{90.6±.9} \\\specialrule{0.05em}{1pt}{1pt}
\multirow{3}{*}{GAT}                                                & w/o Margin               & 98.1±.2          & 95.3±.6          & 88.2±1           & 97.6±.6          & 95.5±.5          & 87.8±.4          \\
& w/o Distill             & 97.7±.8          & 93.4±.5          & 81.2±.4          & 94.8±.7          & 90.2±.5          & 88.6±.7          \\
& MGU                     & \textbf{99.5±.2} & \textbf{98.8±.4} & \textbf{97.7±1}  & \textbf{98.8±.5} & \textbf{98.6±.7} & \textbf{96.8±1}  \\\specialrule{0.05em}{1pt}{1pt}
\multirow{3}{*}{SAGE}                                               & w/o Margin               & 96.4±.9          & 92.5±.9          & 59.6±.8          & 97.7±.2          & 93.6±.8          & 72.6±2           \\
& w/o Distill             & 95.8±1           & 91.9±.7          & 68.3±.4          & 97.8±.5          & 96.5±.5          & 81.4±.7          \\
& MGU                     & \textbf{99.5±.4} & \textbf{98.2±.8} & \textbf{76.9±1}  & \textbf{99.6±.1} & \textbf{97.8±.3} & \textbf{92.4±3} \\\specialrule{0.05em}{1pt}{1pt}
\end{tabular}}}
\label{abl}
\end{table}

\begin{figure}[]
\centering
\scalebox{0.35}{\includegraphics{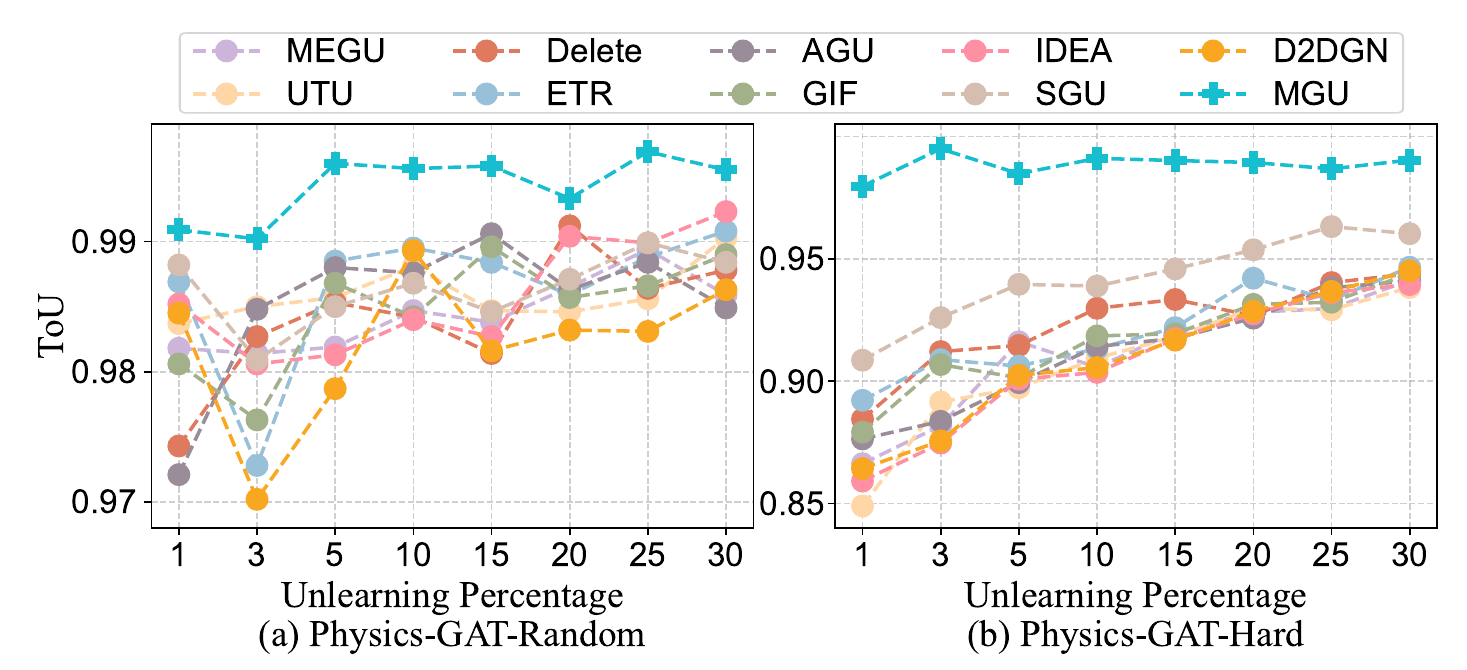}}
\caption{Node unlearning under different ratios.}
\label{scale}
\end{figure}

\subsection{Ablation Study (RQ2)}
We conduct ablation studies to evaluate MGU's key components by constructing two variants: (1) \textbf{w/o Margin} removes the margin-based forgetting module, and (2) \textbf{w/o Distill} excludes the distillation-based preservation module. Table \ref{abl} summarizes the results across three GNN backbones under different difficulty settings. MGU consistently outperforms both variants, with the largest gap in hard tasks: removing the margin module leads to an average 8.5\% drop, while excluding distillation causes a 7.8\% drop. In contrast, the performance gaps on easy and random settings are much smaller (1.2\% and 3.1\%, respectively). These results highlight the importance of both modules, especially for handling hard-to-unlearn cases.

\subsection{Strategy Generalizability (RQ3)}
We evaluate whether our adaptive unlearning strategy (AUS) can enhance existing methods on hard-to-unlearn tasks. Table \ref{improve} reports the results of integrating our strategy (+AUS) into five baseline methods under hard settings. On average, AUS improves performance by 47.8\% on homophily graphs (Cora and Computers) and 44.0\% on the heterophily graph (Chameleon). These results demonstrate the effectiveness and applicability of our strategy across diverse graph structures and baseline methods.

\begin{table}[]
\setlength{\abovecaptionskip}{2pt} 
\setlength{\belowcaptionskip}{0pt} 
\centering
\caption{Baseline performance improvements on hard tasks.}
\resizebox{85mm}{!}{
\setlength{\tabcolsep}{1.2mm}{
\begin{tabular}{c|cc|cc|cc|c}
\specialrule{0.05em}{1pt}{1pt}
\multirow{2}{*}{\textbf{Method}} & \multicolumn{2}{c|}{\textbf{Cora}} & \multicolumn{2}{c|}{\textbf{Computers}} & \multicolumn{2}{c|}{\textbf{Chameleon}} & \multirow{2}{*}{\textbf{Improv}.} \\
   & \textbf{GCN}         & \textbf{GAT}        & \textbf{GCN}           & \textbf{GAT}          & \textbf{SAGE}          & \textbf{FAGCN}         &                          \\\specialrule{0.05em}{1pt}{1pt}
AGU    & 43.4±.3     & 50.5±.4    & 63.6±.8       & 50.5±.4      & 25.8±1        & 36.0±2        & \multirow{2}{*}{61.6\%}  \\
+AUS & \textbf{81.9±.7}     & \textbf{84.2±.6}    & \textbf{83.4±.5}       & \textbf{74.4±.9}      & \textbf{52.3±1}        & \textbf{47.9±3}        &                          \\\specialrule{0.05em}{1pt}{1pt}
D2DGN  & 43.0±2      & 48.1±5     & 62.7±.6       & 61.3±1       & 33.2±1        & 41.4±4        & \multirow{2}{*}{21.5\%}  \\
+AUS & \textbf{51.3±.8}     & \textbf{57.8±3}     & \textbf{68.6±1}        & \textbf{62.8±.3}      & \textbf{55.2±.3}       & \textbf{46.2±.3}       &                          \\\specialrule{0.05em}{1pt}{1pt}
MEGU   & 43.0±3      & 49.4±5     & 63.4±.4       & 81.3±2       & 24.6±2        & 34.7±2        & \multirow{2}{*}{52.4\%}  \\
+AUS & \textbf{79.8±.3}     & \textbf{86.8±.9}    & \textbf{74.1±.7}       & \textbf{83.2±.7}      & \textbf{41.2±3}        & \textbf{57.8±.8}       &                          \\\specialrule{0.05em}{1pt}{1pt}
SGU    & 44.8±1      & 49.0±4     & 63.1±1        & 59.7±2       & 24.8±.7       & 38.3±5        & \multirow{2}{*}{40.4\%}  \\
+AUS & \textbf{56.8±.4}     & \textbf{55.1±3}     & \textbf{78.2±.8}       & \textbf{81.4±1}       & \textbf{52.3±1}        & \textbf{50.6±3}        &                          \\\specialrule{0.05em}{1pt}{1pt}
UTU    & 43.2±2      & 49.7±5     & 63.2±.5       & 57.4±.9      & 25.1±.7       & 36.2±2        & \multirow{2}{*}{50.6\%}  \\
+AUS & \textbf{74.3±.2}     & \textbf{82.3±.3}    & \textbf{72.2±.9}       & \textbf{62.7±3}       & \textbf{53.7±1}        & \textbf{46.5±3}        & \\\specialrule{0.05em}{1pt}{1pt}                        
\end{tabular}}}
\label{improve}
\end{table}

\begin{figure}[]
\centering
\scalebox{0.35}{\includegraphics{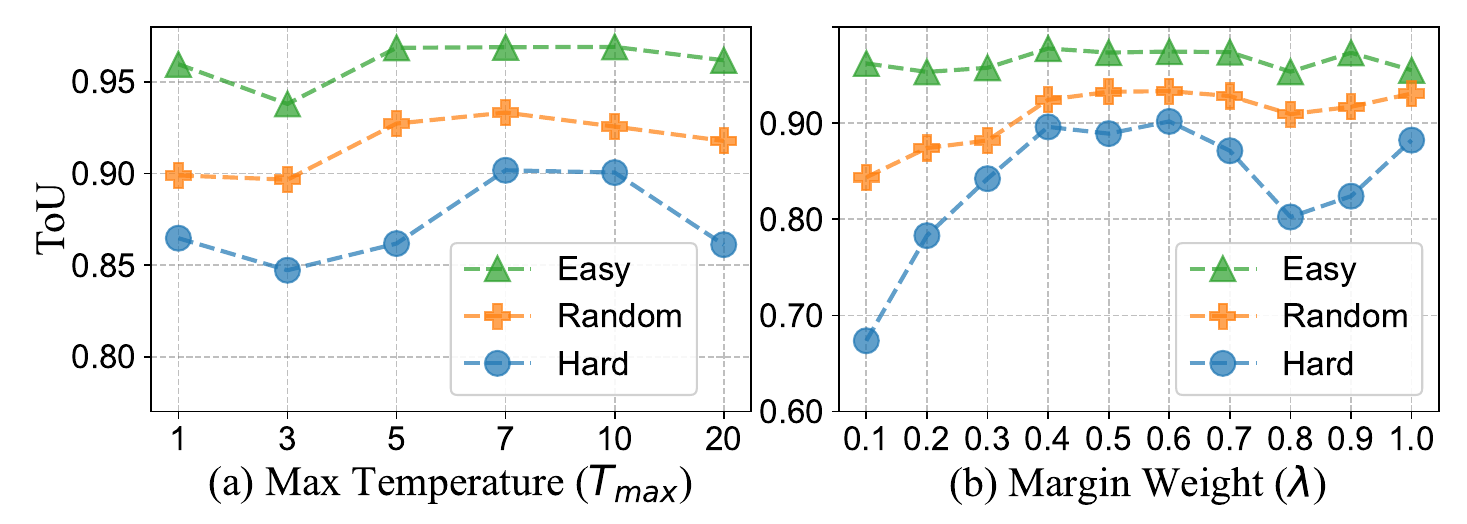}}
\caption{Feature unlearning performance under varying parameters on Photo with GCN.}
\label{para}
\end{figure}

\subsection{Scalability and Parameter Analysis (RQ4)}
We evaluate the performance of MGU under different unlearning scales and parameter settings. Figure \ref{scale} presents ToU results with unlearning ratios from 1\% to 30\%. MGU consistently exhibits superior and stable performance across all scales, demonstrating strong robustness to diverse unlearning requests. In contrast, baseline methods perform significantly worse on hard tasks than on random ones, highlighting the difficulty of unlearning highly memorized elements. Interestingly, as the unlearning ratio increases, the performance gap between unlearned and retrained models narrows, suggesting that deleting more elements removes essential structural information that both models rely on.

Figure \ref{para} presents the sensitivity analysis of two key parameters $T_\text{max}$ and $\lambda$. $T_\text{max}$ controls the maximum softening in distillation. Performance on easy and random tasks remains stable when $T_\text{max}$ ranges from 5 to 10, whereas hard tasks require larger values (7–10) due to their stronger impact on generalization. The margin loss weight $\lambda$ shows a similar trend: performance on easy and random settings remains relatively stable across parameter ranges, while performance on hard settings is more sensitive. Overall, setting $\lambda$ between 0.4 and 0.7 provides robust and consistent performance.

\section{Conclusion}
In this paper, we re-understand graph unlearning from a new perspective of GNN memorization and propose MGU, a memorization-guided unlearning framework that accurately assesses unlearning difficulty and adaptively adjusts unlearning objectives. Extensive experiments demonstrate the superior performance of MGU in forgetting quality, computational efficiency, and utility preservation. Future work will explore memorization mechanisms in diverse graph types and their unlearning implications.


\section*{ACKNOWLEDGMENTS}
This research is supported by ARC Discovery Project DP230100676.

\bibliographystyle{ACM-Reference-Format}
\bibliography{www26}

\end{document}